\documentclass[10pt,twocolumn,letterpaper]{article}

\usepackage{iccv}
\usepackage{times}
\usepackage{epsfig}
\usepackage{graphicx}

\usepackage{ulem}
\usepackage{cuted}
\usepackage{amsfonts} 
\usepackage{amsmath}
\usepackage{array}
\usepackage{mathtools}
\usepackage{comment}
\usepackage{bbm}
\usepackage{bm}
\usepackage{multicol}
\usepackage{multirow}
\usepackage{threeparttable}
\DeclareMathAlphabet{\mathsfit}{\encodingdefault}{\sfdefault}{m}{sl}
\SetMathAlphabet{\mathsfit}{bold}{\encodingdefault}{\sfdefault}{bx}{n}

\usepackage{color}
\usepackage{textcase}
\usepackage{bbding}
\usepackage{array}
\usepackage{booktabs}
\usepackage{enumitem}
\usepackage{arydshln}
\usepackage{makecell}
\usepackage{pifont}
\usepackage{soul}

\makeatletter
\renewcommand*\env@matrix[1][\arraystretch]{%
  \edef\arraystretch{#1}%
  \hskip -\arraycolsep
  \let\@ifnextchar\new@ifnextchar
  \array{*\c@MaxMatrixCols c}}
\makeatother

\iccvfinalcopy 

\def\iccvPaperID{} 

\ificcvfinal\fi

\begin{document}

\title{Graph-Guided MLP-Mixer for Skeleton-Based Human Motion Prediction}

\author{Xinshun Wang\\
Sun Yat-sen University\\
\and
Qiongjie Cui\\
Nanjing University of Science and Technology\\
\and
Chen Chen\\
University of Central Florida\\
\and
Shen Zhao\\
Sun Yat-sen University\\
\and
Mengyuan Liu\\
Peking University\\
}

\maketitle
\ificcvfinal\thispagestyle{empty}\fi

\begin{abstract}

    In recent years, Graph Convolutional Networks (GCNs) have been widely used in human motion prediction, but their performance remains unsatisfactory.
    Recently, MLP-Mixer, initially developed for vision tasks, has been leveraged into human motion prediction as a promising alternative to GCNs, which achieves both better performance and better efficiency than GCNs.
    Unlike GCNs, which can explicitly capture human skeleton's bone-joint structure by representing it as a graph with edges and nodes, MLP-Mixer relies on fully connected layers and thus cannot explicitly model such graph-like structure of human's.
    To break this limitation of MLP-Mixer's, we propose \textit{Graph-Guided Mixer}, a novel approach that equips the original MLP-Mixer architecture with the capability to model graph structure.
    By incorporating graph guidance, our \textit{Graph-Guided Mixer} can effectively capture and utilize the specific connectivity patterns within human skeleton's graph representation.
    In this paper, first we uncover a theoretical connection between MLP-Mixer and GCN that is unexplored in existing research. Building on this theoretical connection, next we present our proposed \textit{Graph-Guided Mixer}, explaining how the original MLP-Mixer architecture is reinvented to incorporate guidance from graph structure.
    Then we conduct an extensive evaluation on the Human3.6M, AMASS, and 3DPW datasets, which shows that our method achieves state-of-the-art performance.
\end{abstract}

\section{Introduction}

\begin{figure}[t]
\centering
\includegraphics[width=0.99\columnwidth]{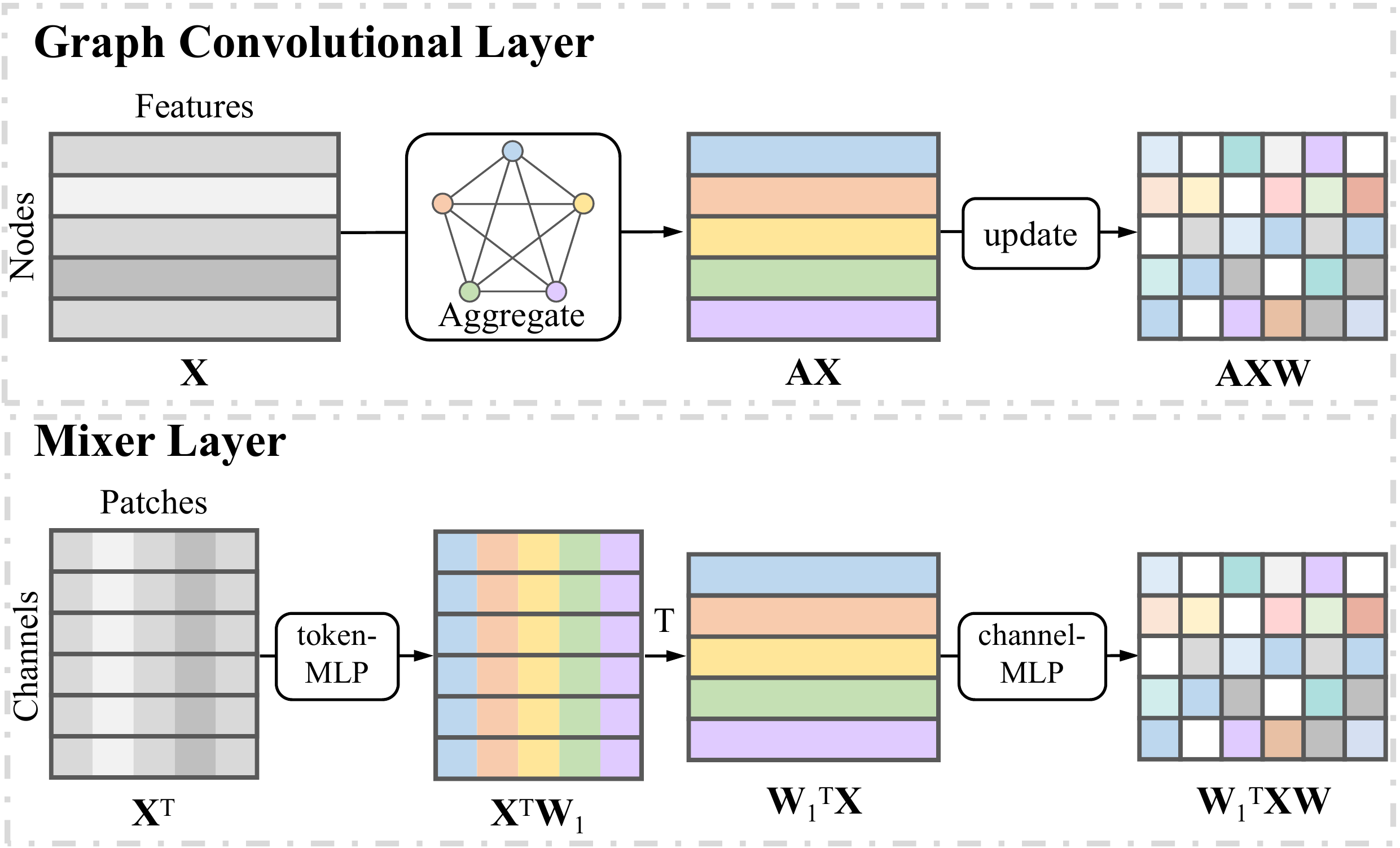} 
\caption{
\textbf{Comparison between MLP-Mixer and GCNs.}
Though differently motivated, MLP-Mixer and GCNs bear a strong analogy in terms of how the individual layer looks.
}
\label{fig1 comparison}
\vspace{-1em}
\end{figure}

    Human motion prediction has much research interest in applications such as virtual human and human-robot interaction \cite{schwind2020effects,unhelkar2018human,von2018recovering}.
    Inspired by various visual analysis techniques \cite{zhang2012imageadmixture,zhang2018detecting,tu2023consistent,tu2022general,tu2023dtcm}, human motion prediction often relies on multimedia techniques such as video, motion capture, and sensor data to extract relevant motion information.
    Currently, Graph Convolutional Networks (GCNs) \cite{kipf2016semi} hold a dominant position in the field \cite{mao2019learning,mao2021multi,ma2022progressively,li2020dynamic,sofianos2021space,chen2021cyclemlp,tang2022image}, derived from the observation that the human body exhibits a graph-like structure with joints as nodes and bones as edges.
    However, one limitation of GCNs lies in their reliance on explicit graph structures, which may oversimplify the complexity of human motion. 
    Moreover, GCNs face challenges in effectively incorporating long-term dependencies and temporal dynamics.
    To overcome these limitations, MLP-Mixer \cite{tolstikhin2021mlp} has recently been leveraged into the field \cite{bouazizi2022motionmixer,guo2023back}, emerging as a promising alternative approach to GCNs with better efficiency and performance.
    MLP-Mixer relies on token mixing and channel mixing operations rather than explicitly incorporating graph structure, and thus is not bound by the constraints of modeling graph-based relationships. This flexibility enables MLP-Mixer to capture fine-grained details, complex interactions and long-term dependencies in human motion by directly operating on tokens representing different aspects of the data.
    MLP-Mixer also comes with its own limitations, the most prominent one of which is that the mixing operation, despite its flexibility, cannot exploit structural information.
    GCNs and MLP-Mixer have each created their own line of research independently of each other, which leads to the question: \ul{Can we combine the advantages of MLP-Mixer and GCN to establish a better approach to human motion prediction?}

            \begin{figure*}[!ht]
            \centering
            \includegraphics[width=0.99\textwidth]{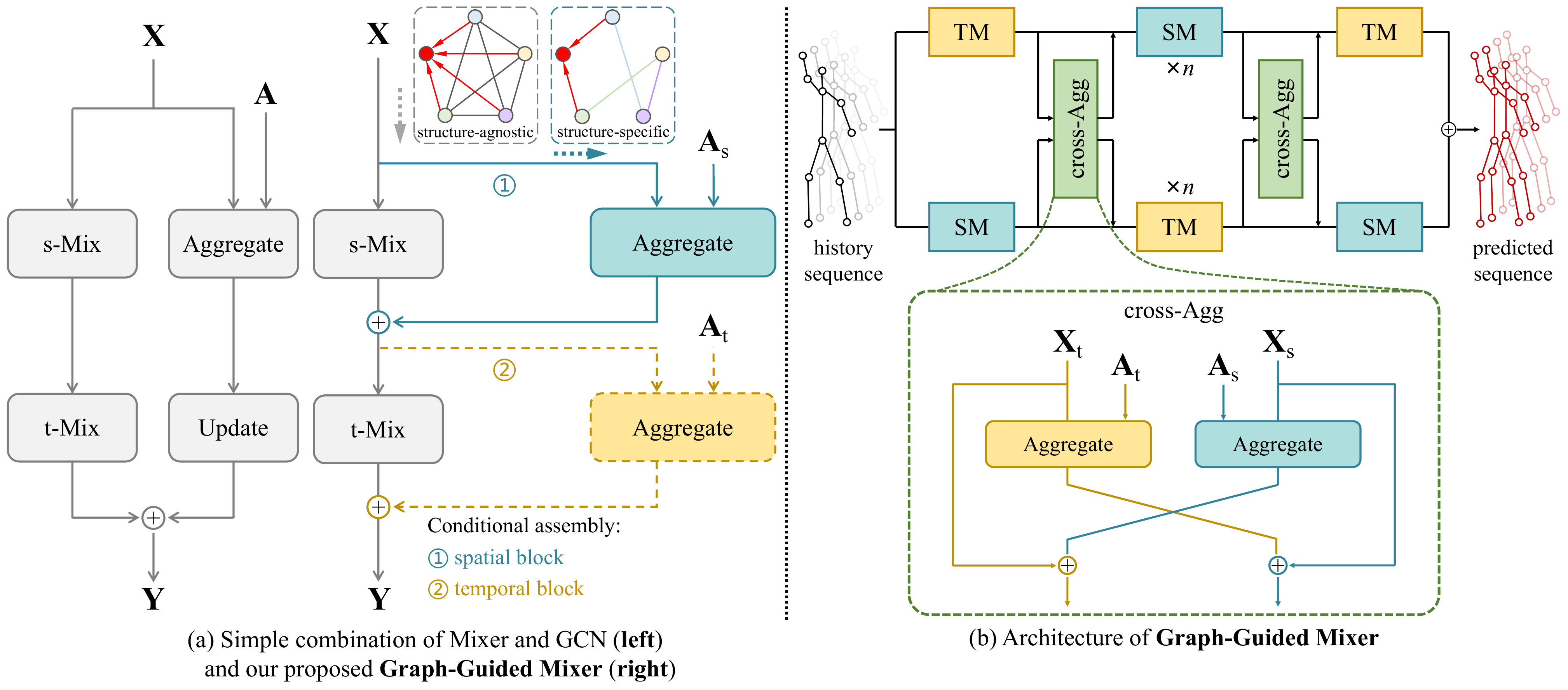} 
            \caption{(Best viewed in color)
            \textbf{Our Graph-Guided Mixer}. (a) Comparison between a simple combination of two models (left) and our proposed Graph-Guided Mixer (right). (b) Architecture of Graph-Guided Mixer.
            Different from a straightforward combination, our Graph-Guided Mixer is developed from the theoretical analogy between Mixers and GCNs.
            }
            \label{network}
            \end{figure*}

    We answer this question by proposing \textit{Graph-Guided Mixer}, which is developed from a theoretical connection that we discover between MLP-Mixer and GCN.
    Consider a simple example where we are given an $N\times D$ matrix $\mathbf{X} = [\mathbf{x}_1,\mathbf{x}_2,\cdots,\mathbf{x}_N]^\top$ of $N$ patches with $D$ features as input.
    Each layer in MLP-Mixer applies channel-mixing and token-mixing, whose overall effect is a combination of matrix products $\mathbf{W}_1^\top\mathbf{X}\mathbf{W}_2$, where $\mathbf{W}_1$ is an $N\times N$ weight matrix and $\mathbf{W}_2$ is a $D\times D$ weight matrix, which are implicitly defined in the channel-mixing MLP and token-mixing MLP respectively.
    The layer in MLP-Mixer bears a strong analogy to the graph convolutional layer in GCNs, which follows an aggregate-update paradigm.
    The aggregation and update also result in a combination of matrix products $\mathbf{A}\mathbf{X}\mathbf{W}$, where $\mathbf{A}$ is an $N\times N$ adjacency matrix for aggregation and $\mathbf{W}$ is a $D\times D$ weight matrix for update.
    This theoretical connection, specifically the finding that the aggregation and channel-mixing both perform a linear combination of node or patch features, makes it possible to inject graph-related information into the original MLP-Mixer architecture with little additional cost. In our proposed \textit{Graph-Guided Mixer}, we use graph structure to guide the MLP-Mixer in a refining way. Continuing from the previous example, our approach is to: add $\mathbf{W}_1^\top$ and $\mathbf{A}$, and then share parameters between $\mathbf{W}_2$ and $\mathbf{W}$.
    This way provides both MLP-Mixer's flexibility and graph's structure-capturing capability.
    To model spatial-temporal dependencies underlying human motion, we design two types of blocks for our \textit{Graph-Guided Mixer}, a spatial and a temporal blocks, which are analogous to the spatial and the temporal graph convolutions respectively.
    
    In summary, our contributions are three-fold:
    \begin{itemize}
        \item We develop a theoretical connection between GCNs and MLP-Mixer, showing that the layer in MLP-Mixer bears a strong analogy to the graph convolutional layer in GCNs, which makes it possible to inject graph guidance into MLP-Mixer with little additional cost.
        \item We propose \textit{Graph-Guided Mixer} based on the theoretical connection, which provides both MLP-Mixer's flexibility and graph's structure-capturing capability.
        \item We conduct extensive experiments, which show that our \textit{Graph-Guided Mixer} outperforms state-of-the-art approaches on Human3.6M, AMASS and 3DPW datasets.
    \end{itemize}

\section{Related Work}
\noindent\textbf{GCN-based human motion prediction.}
    In recent years human motion prediction has been dominated almost exclusively by GCNs. \cite{mao2019learning} is among the first to leverage GCNs in the field, proposing to represent the human skeleton as a learnable fully-connected graph.
    A subsequent work \cite{cui2020learning} introduces a graph with predefined topology based on human skeletal structure. Besides the original skeleton level, human motion can be modeled at different levels using multi-scale graphs \cite{li2020dynamic,dang2021msr,li2021multiscale,zhou2021learning}.
    \cite{li2021multiscale,ma2022progressively} implement the graph convolution layer using two adjacency matrices defined in spatial and temporal domains respectively.
    Motivated by depth-wise separable convolution \cite{chollet2017xception} applied in CNNs, \cite{sofianos2021space} proposes a novel space-time-separable graph convolution.
    \cite{li2021skeleton,li2022skeleton} has investigated the benefits of using spectral GCNs \cite{bruna2013spectral}, which combine graph spectral analysis with graph networks.

\noindent\textbf{Mixer-like models.}
    Our work is also related to MLP-Mixer \cite{tolstikhin2021mlp} and a few Mixer-like models that undertake similar designs.
    In a general sense, the research in computer vision has come full circle back to MLPs, showcased by a prominent line of works
    \cite{tolstikhin2021mlp,touvron2022resmlp,hou2022vision,yu2022s2,guo2022hire,tang2022sparse,tu2022maxim} showing that pure MLP-based networks, when combined with a proper architectural design, achieve competitive performance to CNNs and Transformers \cite{vaswani2017attention} besides being more computationally efficient.
    In MLP-Mixer, each layer mainly involves two steps: token mixing and channel mixing, encoding the spatial information along the flattened spatial dimensions. \cite{hou2022vision} separately encodes the feature representations along the height and width dimensions with linear projections.
    In the spirit to explore non-graph convolutional alternatives, \cite{bouazizi2022motionmixer,guo2023back} employ Mixer-like architectures which deliver state-of-the-art performance besides being more simple and efficient than GCNs. Their models are very much like MLP-Mixers, in which the patch and channel dimensions in MLP-Mixers are branded as spatial and temporal dimensions for motion sequence data.
    Current approaches are either graph convolutional or mixer-like.
    We show in this paper that these two approaches have a theoretical connection to one another that can be exploited to combine their advantages.

\section{The Proposed Method}

Suppose that a motion sequence consists of $T$ consecutive frames, in which each pose has $J$ joints.
The motion sequence is denoted by $\mathbf{S}_{1:T} = [ \mathbf{S}_{1}, \mathbf{S}_{2}, \cdots, \mathbf{S}_{T} ] \in \mathbb{R}^{T\times J\times 3}$.
The task of human motion prediction is to predict the unknown future sequence $\mathbf{S}_{T+1:T+T_\text{f}}$, based on the observed history sequence $\mathbf{S}_{1:T}$.

    \begin{table*}[t]
    \centering
    \resizebox{0.8\textwidth}{!}{
    \begin{tabular}{p{3.5cm}|p{1cm}p{1cm}p{1cm}p{1cm}p{1cm}p{1cm}p{1cm}p{1cm}} \hline
    \multicolumn{9}{c}{Mean Per Joint Position Error (in millimeter)} \\ \hline
    milliseconds & 80 & 160 & 320 & 400 & 560 & 720 & 880 & 1000 \\ \hline
    LTD-50-25 \cite{mao2019learning}	&12.2 	&25.4 	&50.7 	&61.5 	&79.6 	&93.6 	&105.2 	&112.4 	\\
    LTD-10-10 \cite{mao2019learning}	&11.2 	&23.4 	&47.9 	&58.9 	&78.3 	&93.3 	&106.0 	&114.0 	\\
    HisRep \cite{mao2020history}	&10.4 	&22.6 	&47.1 	&58.3 	&77.3 	&91.8 	&104.1 	&112.1 	\\
    MSR-GCN \cite{dang2021msr}	&11.3 	&24.3 	&50.8 	&61.9 	&80.0 	&-	&-	&112.9 	\\
    PGBIG \cite{ma2022progressively}	&10.6 	&23.1 	&47.1 	&57.9 	&76.3 	&90.7 	&102.4 	&109.7 	\\
    siMLPe \cite{guo2023back}	&9.6 	&21.7 	&46.3 	&57.3 	&75.7 	&90.1 	&101.8 	&109.4 	\\ \hline
    Ours	
    &\textcolor{blue}{\textbf{9.4 }}	&\textcolor{blue}{\textbf{21.3 }}	&\textcolor{blue}{\textbf{45.8 }}	&\textcolor{blue}{\textbf{56.7 }}	&\textcolor{blue}{\textbf{75.2 }}	&\textcolor{blue}{\textbf{89.7 }}	&\textcolor{blue}{\textbf{100.9 }}	&\textcolor{blue}{\textbf{108.6 }}
    \\ \hline
    \end{tabular}}
    \resizebox{0.8\textwidth}{!}{
    \begin{tabular}{p{3.5cm}|p{1cm}p{1cm}p{1cm}p{1cm}p{1cm}p{1cm}p{1cm}p{1cm}} \hline
    STSGCN \cite{sofianos2021space}$\dagger$ &10.1&17.1&33.1&38.3&50.8&60.1&68.9&75.6 \\
    GAGCN \cite{zhong2022spatio}$\dagger$ &10.1&16.9&32.5&38.5&50.0&-&-&72.9 \\
    MotionMixer \cite{bouazizi2022motionmixer}$\dagger$ &9.0&13.2&26.9&33.6&46.1&56.5&65.7&71.6 \\ \hline
    Ours$\dagger$ &\textcolor{blue}{\textbf{6.9}}     &\textcolor{blue}{\textbf{13.0}}  &\textcolor{blue}{\textbf{24.9}}  &\textcolor{blue}{\textbf{30.8}} &\textcolor{blue}{\textbf{41.6}} &\textcolor{blue}{\textbf{51.4}} &\textcolor{blue}{\textbf{59.7}} &\textcolor{blue}{\textbf{65.0} }\\ \hline
    \end{tabular}}
    \vspace{0.5em}
    \caption{\textbf{Results on H3.6M} at different prediction time steps in terms of mean per joint position error (MPJPE) in millimeter. 256 samples are tested for each action. $\dagger$ indicates methods that compute the average error over all frames, whose results are taken from the paper \cite{bouazizi2022motionmixer}. Otherwise, the methods are evaluated at each particular frame, whose results are taken from the paper \cite{guo2023back}. We can see that our approach consistently outperforms other approaches under both evaluation protocols.
    }
    \label{h36m}
    \end{table*}

\subsection{Graph-Guided Mixer}\label{our method}

\noindent\textbf{Theoretical connection.}
Given a motion sequence of size $T\times J\times 3$, we first reshape it into a sequence of flattened poses.
This results in a matrix $\mathbf{X} = \left[ \mathbf{x}_1,\mathbf{x}_2,\cdots,\mathbf{x}_N \right] ^\top \in\mathbb{R}^{N\times T}$, with $N=3J$. Each row vector $\mathbf{x}_{n}^\top$ corresponds to the trajectory of a joint coordinate, and each column vector a flattened pose.

In Mixer-based methods, the network employs MLPs to perform spatial mixing and temporal mixing along the two dimensions of $\mathbf{X}$ respectively.
In classic GCN-based methods, $\mathbf{X}$ is treated as a graph with $N$ nodes and $T$ features for each node. Then the network performs graph convolutions over the graph based on trainable adjacencies.
A mixer layer and a graph convolutional layer can both be implemented based on matrix multiplications.
In each mixer layer, the channel-mixing transposes the input, right-multiplies it with a weight matrix $\mathbf{W}_1$ and then transposes back the product.
The token-mixing then right-multiplies it with another weight matrix $\mathbf{W}_2$.
These two mixing operations result in a combination of matrix products in the following form:
\begin{equation}\label{mixer}
    f_\text{Mix} (\mathbf{X}) = \mathbf{W}_1^\top\mathbf{X}\mathbf{W}_2,
\end{equation}
where $\mathbf{W}_1\in\mathbb{R}^{N\times N}$ and $\mathbf{W}_2\in\mathbb{R}^{T\times T}$ are both trainable weight matrices implicitly given by the channel-mixing and token-mixing MLPs respectively.
This function shows that the overall effect of a mixer layer can be seen as to left-multiply the input with a weight matrix then right-multiply it with another.
Since the weight matrices are set as trainable parameters, the transpose of them does not make much of a difference on the practical side.
The mixer layer bears a strong analogy to the graph convolutional layer in standard GCNs, which follows an aggregate-update paradigm.
The aggregation is left-multiplying the input with an adjacency matrix, and the update is right-multiplying it with a weight matrix. The overall action of a graph convolutional layer is:
\begin{equation}\label{graph conv}
    f_\text{GraphConv} (\mathbf{X}) = \mathbf{A}\mathbf{X}\mathbf{W},
\end{equation}
where $\mathbf{A}\in\mathbb{R}^{N\times N}$ is the adjacency matrix and $\mathbf{W}\in\mathbb{R}^{T\times T}$ is a trainable weight usually set as a square matrix.

\noindent\textbf{Spatial and temporal mixing.}
The mixing blocks adopt a similar design to those in \cite{tolstikhin2021mlp,guo2023back} which involve fully-connected layers, transpose operations and layer normalization. The spatial mixing (s-Mix) and temporal mixing (t-Mix) are defined using matrix operations as:
\begin{equation}\label{basic mixing}
\begin{split}
    &\mathbf{U} = \mathcal{S} (\mathbf{X}; \Theta_1) = (  \mathbf{X}^\top\mathbf{W}_{\theta1} )^\top ;\\
    &\mathbf{Y} = \mathcal{T} (\mathbf{U}; \Theta_2) =  \mathbf{U}\mathbf{W}_{\theta2} ,
\end{split}
\end{equation}
where $\mathcal{S}(\cdot; \Theta_1)$ is the spatial mixing function with a trainable weight matrix $\mathbf{W}_{\theta1}\in\mathbb{R}^{N\times N}$ parameterized by a set of parameters $\Theta_1$, $\mathcal{T}(\cdot; \Theta_2)$ is the temporal mixing function with a trainable weight matrix $\mathbf{W}_{\theta2}\in\mathbb{R}^{T\times T}$ parameterized by a set of parameters $\Theta_2$.

\noindent\textbf{Graph-guided mixing.}
The \textit{Graph-Guided Mixer} is constructed with different flavors of assembly, resulting in a spatial or a temporal block.
The conditional assembly is based on the fact that matrix multiplications are associative, so the order in which temporal and spatial mixing or aggregation and update are applied does not change the layer outcome.
For the spatial block, its overall operation $\mathcal{M}_\text{s}(\mathbf{X})$ is given by:
\begin{equation}\label{spatial meta mixer}
    \begin{split}
        &\mathbf{U} = \text{LN} \big(  \mathcal{S} (\mathbf{X}; \Theta_1) + \mathcal{Z}(\mathbf{A}_\text{s}, \mathbf{X}; \Theta_\text{s})   \big) ;\\
        &\mathbf{Y} = \text{LN} \big( \mathcal{T} (\mathbf{U}; \Theta_2) \big),
    \end{split}
\end{equation}
where $\mathbf{A}_\text{s} \in\mathbb{R}^{N\times N}$ is a spatial adjacency matrix, specified by the skeleton's joint-bone structure, and LN is layer normalization.

The operation of the temporal block $\mathcal{M}_\text{t}(\mathbf{X})$ is given by:
\begin{equation}\label{temporal meta mixer}
    \begin{split}
        &\mathbf{U} = \text{LN} \big(  \mathcal{S} (\mathbf{X}; \Theta_1) \big);\\
        &\mathbf{Y} = \text{LN} \big(     \mathcal{T} (\mathbf{U}; \Theta_2) +    \mathcal{Z}(\mathbf{A}_\text{t}, \mathbf{U}; \Theta_\text{t})         \big),
    \end{split}
\end{equation}
where $\mathbf{A}_\text{t} \in\mathbb{R}^{D\times D}$ is a temporal adjacency matrix, which is set as a tridiagonal matrix. The idea behind such a setting is to limit the neighborhood of each node across from its last frame to its next frame of the same joint.

\begin{table*}[!ht]
    \centering
    \resizebox{\textwidth}{!}{
    \begin{tabular}{c|cccc|cccc|cccc|cccc} \hline
    scenarios    & \multicolumn{4}{c|}{Walking} & \multicolumn{4}{c|}{Eating} &  \multicolumn{4}{c|}{Smoking} & \multicolumn{4}{c}{Discussion} \\ \hline
    milliseconds & 80 & 400 & 560 & 1000 & 80 & 400 & 560 & 1000 & 80 & 400 & 560 & 1000 & 80 & 400 & 560 & 1000 \\ \hline
    LTD-50-25 \cite{mao2019learning}	&12.3 	&44.4 	&50.7 	&60.3 	&7.8 	&38.6 	&51.5 	&75.8 	&8.2 	&39.5 	&50.5 	&72.1 	&11.9 	&68.1 	&88.9 	&118.5 	\\
    LTD-10-10 \cite{mao2019learning}	&11.1 	&42.9 	&53.1 	&70.7 	&7.0 	&37.3 	&51.1 	&78.6 	&7.5 	&37.5 	&49.4 	&71.8 	&10.8 	&65.8 	&88.1 	&121.6 	\\
    HisRep\cite{mao2020history}	&10.0 	&39.8 	&47.4 	&58.1 	&6.4 	&36.2 	&50.0 	&75.7 	&7.0 	&36.4 	&47.6 	&69.5 	&10.2 	&65.4 	&86.6 	&119.8 	\\
    MSR-GCN\cite{dang2021msr}	&10.8 	&42.4 	&53.3 	&63.7 	&6.9 	&36.0 	&50.8 	&75.4 	&7.5 	&37.5 	&50.5 	&72.1 	&10.4 	&65.0 	&87.0 	&116.8 	\\
    PGBIG \cite{ma2022progressively}	&11.2 	&42.8 	&49.6 	&58.9 	&6.5 	&36.8 	&50.0 	&74.9 	&7.3 	&37.5 	&48.8 	&69.9 	&10.2 	&64.4 	&86.1 	&116.9 	\\
    siMLPe \cite{guo2023back}	&9.9 	&39.6 	&46.8 	&55.7 	&5.9 	&36.1 	&49.6 	&74.5 	&6.5 	&36.3 	&47.2 	&69.3 	&9.4 	&64.3 	&85.7 	&116.3 	\\ \hline
    Ours	&\textcolor{blue}{\textbf{9.3}} &\textcolor{blue}{\textbf{38.6}} &\textcolor{blue}{\textbf{45.9}} &\textcolor{blue}{\textbf{55.6}} &\textcolor{blue}{\textbf{5.6}} &\textcolor{blue}{\textbf{34.1}} &\textcolor{blue}{\textbf{48.6}} &\textcolor{blue}{\textbf{74.3}} &\textcolor{blue}{\textbf{6.3}} &\textcolor{blue}{\textbf{36.3}} &\textcolor{blue}{\textbf{46.8}} &\textcolor{blue}{\textbf{68.8}} &\textcolor{blue}{\textbf{9.2}} &\textcolor{blue}{\textbf{63.4}} &\textcolor{blue}{\textbf{84.3}} &\textcolor{blue}{\textbf{115.0}}
     \\
    \hline
    \end{tabular}}
    \resizebox{\textwidth}{!}{
    \begin{tabular}{c|cccc|cccc|cccc|cccc} \hline
    scenarios    & \multicolumn{4}{c|}{Directions} & \multicolumn{4}{c|}{Greeting} &  \multicolumn{4}{c|}{Phoning} & \multicolumn{4}{c}{Posing} \\ \hline
    milliseconds & 80 & 400 & 560 & 1000 & 80 & 400 & 560 & 1000 & 80 & 400 & 560 & 1000 & 80 & 400 & 560 & 1000 \\ \hline
    LTD-50-25 \cite{mao2019learning}	&8.8 	&58.0 	&74.2 	&105.5 	&16.2 	&82.6 	&104.8 	&136.8 	&9.8 	&50.8 	&68.8 	&105.1 	&12.2 	&79.9 	&110.2 	&174.8 	\\
    LTD-10-10 \cite{mao2019learning}	&8.0 	&54.9 	&76.1 	&108.8 	&14.8 	&79.7 	&104.3 	&140.2 	&9.3 	&49.7 	&68.7 	&105.1 	&10.9 	&75.9 	&109.9 	&171.7 	\\
    HisRep\cite{mao2020history}	&7.4 	&56.5 	&73.9 	&106.5 	&13.7 	&78.1 	&101.9 	&138.8 	&8.6 	&49.2 	&67.4 	&105.0 	&10.2 	&75.8 	&107.6 	&178.2 	\\
    MSR-GCN\cite{dang2021msr}	&7.7 	&56.2 	&75.8 	&105.9 	&15.1 	&85.4 	&106.3 	&136.3 	&9.1 	&49.8 	&67.9 	&104.7 	&10.3 	&75.9 	&112.5 	&176.5 	\\
    PGBIG \cite{ma2022progressively}	&7.5 	&56.0 	&73.3 	&105.9 	&14.0 	&77.3 	&100.2 	&136.4 	&8.7 	&48.8 	&66.5 	&102.7 	&10.2 	&73.3 	&102.8 	&167.0 	\\
    siMLPe \cite{guo2023back}	&6.5 	&55.8 	&73.1 	&106.7 	&12.4 	&77.3 	&99.8 	&137.5 	&8.1 	&48.6 	&66.3 	&103.3 	&8.8 	&73.8 	&103.4 	&168.7 	\\\hline
    
    Ours	&\textcolor{blue}{\textbf{5.9}} &\textcolor{blue}{\textbf{55.7}} &\textcolor{blue}{\textbf{72.1}} &106.8 &\textcolor{blue}{\textbf{12.4}} &\textcolor{blue}{\textbf{76.2}} &\textcolor{blue}{\textbf{99.4}} &\textcolor{blue}{\textbf{134.5}} &\textcolor{blue}{\textbf{7.7}} &\textcolor{blue}{\textbf{47.8}} &\textcolor{blue}{\textbf{66.0}} &103.1 &\textcolor{blue}{\textbf{8.8}} &\textcolor{blue}{\textbf{72.5}} &\textcolor{blue}{\textbf{100.7}} &\textcolor{blue}{\textbf{165.3}} 	
    \\
    \hline
    \end{tabular}}
    \resizebox{\textwidth}{!}{
    \begin{tabular}{c|cccc|cccc|cccc|cccc} \hline
    scenarios    & \multicolumn{4}{c|}{Purchases} & \multicolumn{4}{c|}{Sitting} &  \multicolumn{4}{c|}{Sitting Down} & \multicolumn{4}{c}{Taking Photo} \\ \hline
    milliseconds & 80 & 400 & 560 & 1000 & 80 & 400 & 560 & 1000 & 80 & 400 & 560 & 1000 & 80 & 400 & 560 & 1000 \\ \hline
    LTD-50-25 \cite{mao2019learning}	&15.2 	&78.1 	&99.2 	&134.9 	&10.4 	&58.3 	&79.2 	&118.7 	&17.1 	&76.4 	&100.2 	&143.8 	&9.6 	&54.3 	&75.3 	&118.8 	\\
    LTD-10-10 \cite{mao2019learning}	&13.9 	&75.9 	&99.4 	&135.9 	&9.8 	&55.9 	&78.5 	&118.8 	&15.6 	&71.7 	&96.2 	&142.2 	&8.9 	&51.7 	&72.5 	&116.3 	\\
    HisRep\cite{mao2020history}	&13.0 	&73.9 	&95.6 	&134.2 	&9.3 	&56.0 	&76.4 	&115.9 	&14.9 	&72.0 	&97.0 	&143.6 	&8.3 	&51.5 	&72.1 	&115.9 	\\
    MSR-GCN\cite{dang2021msr}	&13.3 	&77.8 	&99.2 	&134.5 	&9.8 	&55.5 	&77.6 	&115.9 	&15.4 	&73.8 	&102.4 	&149.4 	&8.9 	&54.4 	&77.7 	&121.9 	\\
    PGBIG \cite{ma2022progressively}	&13.2 	&74.0 	&95.7 	&132.1 	&9.1 	&54.6 	&75.1 	&114.8 	&14.7 	&70.0 	&94.4 	&139.0 	&8.2 	&50.2 	&70.5 	&112.9 	\\
    siMLPe \cite{guo2023back}	&11.7 	&72.4 	&93.8 	&132.5 	&8.6 	&55.2 	&75.4 	&114.1 	&13.6 	&70.8 	&95.7 	&142.4 	&7.8 	&50.8 	&71.0 	&112.8 	\\ \hline
    Ours	&\textcolor{blue}{\textbf{10.5}} &\textcolor{blue}{\textbf{70.7}} &\textcolor{blue}{\textbf{92.4}} &\textcolor{blue}{\textbf{129.1}} &\textcolor{blue}{\textbf{8.1}} &\textcolor{blue}{\textbf{53.8}} &\textcolor{blue}{\textbf{74.8}} &\textcolor{blue}{\textbf{113.8}} &13.7 &\textcolor{blue}{\textbf{69.2}} &95.6 &\textcolor{blue}{\textbf{138.6}} &\textcolor{blue}{\textbf{7.8}} &\textcolor{blue}{\textbf{49.8}} &72.1 &114.3 	\\
    \hline
    \end{tabular}}
    \resizebox{\textwidth}{!}{
    \begin{tabular}{c|cccc|cccc|cccc|cccc} \hline
    scenarios    & \multicolumn{4}{c|}{Waiting} & \multicolumn{4}{c|}{Walking Dog} &  \multicolumn{4}{c|}{Walking Together} & \multicolumn{4}{c}{Average} \\ \hline
    milliseconds & 80 & 400 & 560 & 1000 & 80 & 400 & 560 & 1000 & 80 & 400 & 560 & 1000 & 80 & 400 & 560 & 1000 \\ \hline
    LTD-50-25 \cite{mao2019learning}	&10.4 	&59.2 	&77.2 	&108.3 	&22.8 	&88.7 	&107.8 	&156.4 	&10.3 	&46.3 	&56.0 	&65.7 	&12.2 	&61.5 	&79.6 	&112.4 	\\
    LTD-10-10 \cite{mao2019learning}	&9.2 	&54.4 	&73.4 	&107.5 	&20.9 	&86.6 	&109.7 	&150.1 	&9.6 	&44.0 	&55.7 	&69.8 	&11.2 	&58.9 	&78.3 	&114.0 	\\
    HisRep\cite{mao2020history}	&8.7 	&54.9 	&74.5 	&108.2 	&20.1 	&86.3 	&108.2 	&146.9 	&8.9 	&41.9 	&52.7 	&64.9 	&10.4 	&58.3 	&77.3 	&112.1 	\\
    MSR-GCN\cite{dang2021msr}	&10.4 	&62.4 	&74.8 	&105.5 	&24.9 	&112.9 	&107.7 	&145.7 	&9.2 	&43.2 	&56.2 	&69.5 	&11.3 	&61.9 	&80.0 	&112.9 	\\
    PGBIG \cite{ma2022progressively}	&8.7 	&53.6 	&71.6 	&103.7 	&20.4 	&84.6 	&105.7 	&145.9 	&8.9 	&43.8 	&54.4 	&64.6 	&10.6 	&57.9 	&76.3 	&109.7 	\\
    siMLPe \cite{guo2023back}	&7.8 	&53.2 	&71.6 	&104.6 	&18.2 	&83.6 	&105.6 	&141.2 	&8.4 	&41.2 	&50.8 	&61.5 	&9.6 	&57.3 	&75.7 	&109.4 	\\\hline
    Ours  &\textcolor{blue}{\textbf{7.6}} &\textcolor{blue}{\textbf{51.8}} &73.0 &\textcolor{blue}{\textbf{102.7}} &19.0 &84.2 &\textcolor{blue}{\textbf{105.1}} &\textcolor{blue}{\textbf{140.4}} &8.5 &\textcolor{blue}{\textbf{40.9}} &52.4 &63.6 &\textcolor{blue}{\textbf{9.4}} &\textcolor{blue}{\textbf{56.7}} &\textcolor{blue}{\textbf{75.2}} &\textcolor{blue}{\textbf{108.6}}    \\
    \hline
    \end{tabular}}
    \vspace{0.2em}
    \caption{\textbf{Action-wise results on H3.6M} at different prediction time steps in terms of MPJPE in millimeter. 256 samples are tested for each action. The methods are evaluated at each particular frame, whose results are taken from the paper \cite{guo2023back}. Our model achieves state-of-the-art performance on nearly all actions and time steps, especially as the prediction time grows.
    }
    \label{h36m per action}
    \vspace{-1em}
    \end{table*}

\subsection{Network Architecture}

The spatial and temporal blocks serve as the building blocks of the overall network, termed \textit{Graph-Guided Mixer}.
Figure \ref{network} depicts the overall architecture of the proposed network.
The network adopts a two-branch design assembled in a symmetrical fashion.
In each branch, we employ a number $n+2$ of blocks in successive order. The first and the last blocks are differently assembled from the blocks in the middle.
Aside from a series of blocks in each branch, we employ two fusion blocks between branches after the first blocks and before the last blocks respectively. 
The fusion blocks are implemented using aggregation to allow for the exchange of spatio-temporal features between branches.
The outputs of two branches are added together to produce the final output.
Following \cite{guo2023back}, other components of the network include Discrete Cosine Transform (DCT), residual connection and data augmentation, which are consistent with theirs.

    \begin{table*}[t]
    \centering
    \resizebox{\textwidth}{!}{
    \begin{tabular}{c|cccccccc|cccccccc} \hline
    Dataset & \multicolumn{8}{c|}{AMASS-BMLrub} & \multicolumn{8}{c}{3DPW} \\ \hline
    milliseconds & 80 & 160 & 320 & 400 & 560 & 720 & 880 & 1000 & 80 & 160 & 320 & 400 & 560 & 720 & 880 & 1000 \\ \hline
    convSeq2Seq \cite{li2018convolutional}	&20.6 	&36.9 	&59.7 	&67.6 	&79.0 	&87.0 	&91.5 	&93.5 	&18.8 	&32.9 	&52.0 	&58.8 	&69.4 	&77.0 	&83.6 	&87.8 	\\
    LTD-10-10 \cite{mao2019learning}	&10.3 	&19.3 	&36.6 	&44.6 	&61.5 	&75.9 	&86.2 	&91.2 	&12.0 	&22.0 	&38.9 	&46.2 	&59.1 	&69.1 	&76.5 	&81.1 	\\
    LTD-10-25 \cite{mao2019learning}	&11.0 	&20.7 	&37.8 	&45.3 	&57.2 	&65.7 	&71.3 	&75.2 	&12.6 	&23.2 	&39.7 	&46.6 	&57.9 	&65.8 	&71.5 	&75.5 	\\
    HisRep \cite{mao2020history}	&11.3 	&20.7 	&35.7 	&42.0 	&51.7 	&58.6 	&63.4 	&67.2 	&12.6 	&23.1 	&39.0 	&45.4 	&56.0 	&63.6 	&69.7 	&73.7 	\\
    siMLPe \cite{guo2023back} &10.8 	&19.6 	&34.3 	&40.5 	&50.5 	&57.3 	&62.4 	&65.7 	&12.1 	&22.1 	&38.1 	&44.5 	&54.9 	&62.4 	&68.2 	&72.2 	\\ \hline
    Ours
    &\textcolor{blue}{\textbf{10.4 }}	&\textcolor{blue}{\textbf{19.1 }}	&38.9 	&\textcolor{blue}{\textbf{40.2 }}	&\textcolor{blue}{\textbf{50.1 }}	&\textcolor{blue}{\textbf{56.6 }}	&\textcolor{blue}{\textbf{61.9 }}	&\textcolor{blue}{\textbf{65.4 }}	&\textcolor{blue}{\textbf{12.0 }}	&\textcolor{blue}{\textbf{22.1} }	&\textcolor{blue}{\textbf{37.8 }}	&\textcolor{blue}{\textbf{44.1 }}	&\textcolor{blue}{\textbf{54.5 }}	&\textcolor{blue}{\textbf{62.1 }}	&\textcolor{blue}{\textbf{68.0 }} & 72.3
    \\ \hline
    \end{tabular}}
    \vspace{0.2em}
    \caption{\textbf{Results on AMASS and 3DPW} at different prediction time steps. The results are taken from the paper \cite{guo2023back}. 
    }
    \label{amass 3dpw}
    \vspace{-1em}
    \end{table*}

    \begin{figure*}[ht]
    \centering
    \includegraphics[width=0.95\textwidth]{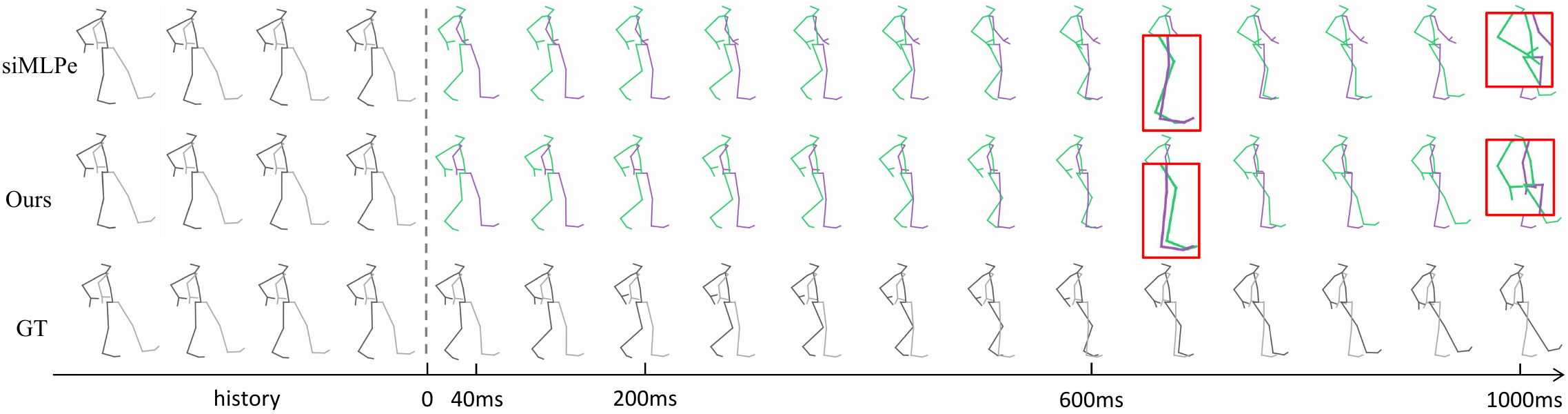} 
    \vspace{-0.5em}
    \caption{
    \textbf{Qualitative results on H3.6M.} 
    We visualize a walking sample of predicted pose sequences and their ground truths.
    }
    \label{viz}
    \end{figure*}

\section{Experiment}

\subsection{Datasets}
\noindent
\textbf{Human3.6M (H3.6M). }
H3.6M \cite{ionescu2013human3} is a large-scale dataset for human-related tasks. It involves 15 types of actions performed by 7 actors. Following previous work, We preprocess and convert the data to 3D coordinates. Each pose is characterized by 22 joints. We use subject 11 and subject 5 for validation and testing respectively, and the remaining 5 subjects for training.

\noindent
\textbf{AMASS. }
AMASS \cite{mahmood2019amass} is a collection of multiple Mocap datasets unified by SMPL parameterization. Following \cite{guo2023back}, we use AMASS-BMLrub as the test set and split the rest of the AMASS dataset into training and validation sets.

\noindent
\textbf{3D Pose in the Wild (3DPW). }
3DPW \cite{von2018recovering} covers activities captured from both indoor and outdoor scenes. Following \cite{guo2023back} we evaluate 18 joints using the model trained on AMASS.

\subsection{Experimental Setup}
\noindent
\textbf{Evaluation Metrics. }
We report the experimental results in terms of the Mean Per Joint Position Error (MPJPE), which is the most preferred metric in the community.
The lower the MPJPE, the better the performance.

\noindent
\textbf{Implementation Details. }
We implemented our network using Pytorch. We trained the model on NVIDIA RTX 3080 Ti GPU for 60k iterations with a batch size of 256. We used Adam optimizer \cite{kingma2014adam} with learning rate starting at 0.0005 and dropping to 0.000005 after 55k iterations.

\subsection{Comparison with State-of-the-Art Methods}
In this subsection we report quantitative results of state-of-the-art methods and ours for motion prediction on H3.6M, AMASS and 3DPW. We also present qualitative evaluation by visualizing predicted samples.

\noindent
\textbf{Comparison on H3.6M. }
The H3.6M dataset is the currently most preferred dataset for evaluating the performance of human motion prediction methods. We first present the quantitative results on H3.6M with the utmost detail and precision. We also provide qualitative results of different methods by visualizing some predicted samples in H3.6M.
Following \cite{guo2023back}, we test 256 samples for each action. The baseline methods follow two different testing protocols, one as in \cite{mao2020history} where the errors are computed at each time step, and the other as in \cite{sofianos2021space} where the errors are computed by averaging all time steps.
In fair comparison, we evaluate our method under both of the two testing protocols. The two protocols are differentiated by a $\dagger$ sign.
In Table \ref{h36m}, we compare our methods with other state-of-the-art methods in terms of average MPJPEs for all actions on average at different time steps.
Our method achieves the best results under both testing protocols.
For more comparison, we also report the action-wise results of all 15 actions in Table \ref{h36m per action}.

\noindent
\textbf{Comparison on AMASS and 3DPW. }
We report the results on AMASS and 3DPW in Table \ref{amass 3dpw}. Following \cite{guo2023back}, the models are trained on AMASS and tested on AMASS-BMLrub and 3DPW. There are two different evaluation protocols used by different methods in the field, following \cite{sofianos2021space} and \cite{mao2019learning} respectively. We evaluate our method under both two evaluation protocols.

\subsection{Ablation Study}
In this subsection, we conduct a series of ablation experiments on the essential components of our network to analyze the advantages of our method.

\noindent\textbf{Importance of Meta-Mixing.}\label{paragraph ablation setting meta-mixing}
We first show that developing the theoretical connection between GCNs and Mixers is important and necessary for the task, which is the theoretical motivation of our proposed Graph-Guided Mixer.
We demonstrate this argument in a progressive way. First, we build a baseline model which is a simple combination of GCN and Mixer with a two-stream design, whose outputs are added at the end of each layer. 
Then we gradually incorporate each essential component into this simple model, with the following assembly choices:
\textbf{a)} share parameters between the update and temporal mixing;
\textbf{b)} add aggregation and spatial mixing;
\textbf{c)} share parameters between the aggregation and spatial mixing;
\textbf{d)} add update and temporal mixing;
\textbf{e)} construct two parallel branches with symmetric assembly to one another;
\textbf{f)} adding fusion blocks across two branches.
The results in Table \ref{ablation meta-mixing} verify the importance and necessity of unifying Mixers and GCNs into a general approach rather than simply combining them.
It also shows the advantages brought by techniques used in our network such as parameter tying.

\noindent\textbf{Ablation on network architecture.}
Next we ablate on different design choices for the network.
First, we assemble the network such that each branch has the same type of Graph-Guided Mixer blocks, instead of interleaved spatial and temporal blocks in each branch.
Second, we apply fusion connections after each block.
Third, we use the same type of blocks to assemble the entire network.
The results under each of these settings are shown in Table \ref{ablation network architecture}, which verifies the effectiveness of our network architecture.

    \begin{table}[t]
    \centering
    \resizebox{\columnwidth}{!}{
    \begin{tabular}{m{2.7cm}<{\raggedright}|cccccc} \hline
    Setting & 80 & 160 & 320 & 400 & 560 & 1000 \\ \hline
    baseline    &10.1& 22.5& 47.4& 58.5& 78.3& 115.0 \\
    a           &10.2& 23.0& 48.1& 59.1& 78.0& 113.2 \\
    a,b         &10.2 &22.6 &47.7 &59.0 &77.7 &112.2 \\
    c           &11.2& 25.1& 51.5& 62.6& 81.2& 114.4\\
    c,d         &10.1& 22.7& 47.6& 58.9& 77.5&  112.5 \\
    a,b,c,d        &9.7& 21.7& 45.9& 56.8& 75.4& 109.3 \\
    a,b,c,d,e        &9.6& 21.7& 45.8& 56.8& 75.3&109.0 \\ \hline
    a,b,c,d,e,f (default) & \textbf{9.4} & \textbf{21.3} & \textbf{45.8} & \textbf{56.7} & \textbf{75.2} & \textbf{108.6} \\ \hline
    \end{tabular}}
    \vspace{0.2em}
    \caption{\textbf{Ablation on Graph-Guided Mixer}.
    Each group of settings labeled a--f (see ``Importance of Graph-Guided mixing'') results in a different model that is used to evaluate the critical components of Graph-Guided Mixer.
    }
    \label{ablation meta-mixing}
    \end{table}

    \begin{table}[t]
    \centering
    \resizebox{\columnwidth}{!}{
    \begin{tabular}{m{2.5cm}<{\raggedright}|cccccc} \hline
    Design choice & 80 & 160 & 320 & 400 & 560 & 1000 \\ \hline
    w/o interleave    &9.7& 22.1& 46.8& 57.9& 76.8& 110.3 \\
    full fuse          &10.4& 23.0& 47.6& 58.6& 77.6& 112.4 \\
    all SM         &9.7& 21.8& 46.2& 57.1& 75.8& 109.7 \\
    all TM         &10.2& 22.6& 47.2& 58.3& 77.0& 111.0 \\ \hline
    Default         & \textbf{9.4} & \textbf{21.3} & \textbf{45.8} & \textbf{56.7} & \textbf{75.2} & \textbf{108.6} \\ \hline
    \end{tabular}}
    \vspace{0.2em}
    \caption{\textbf{Ablation on network architecture}.
    ``w/o interleave'' means each branch has the same type of blocks instead of interleaved ones. ``full fuse'' means applying fusion connections after each block. ``all SM'' and ``all TM'' means using the same type of blocks to assemble the entire network.
    }
    \label{ablation network architecture}
    \end{table}

    \begin{table}[t]
    \centering
    \resizebox{0.95\columnwidth}{!}{
    \begin{tabular}{m{1.8cm}|c|c|m{0.8cm}<{\centering}m{0.8cm}<{\centering}m{0.8cm}<{\centering}} \hline
    Nb. Blocks & Param. (M) & FLOPs & 80  & 560 & 1000 \\ \hline
    12  & 0.26           &1.4G             &10.1& 77.5& 113.2   \\
    24  & 0.50               &2.5G            &9.9& 76.4& 111.7  \\
    48 (default) & 0.96      &4.9G            & \textbf{9.4}  & \textbf{75.2} & \textbf{108.6} \\
    64  & 1.30                &6.5G      &9.5& 75.2& 109.9  \\
    96  & 1.95        &9.6G            &9.7& 75.4& 110.2  \\ \hline
    \end{tabular}}
    \vspace{0.2em}
    \caption{\textbf{Ablation on block number}.
    The best result is achieved with 48 blocks with less than one million parameters.
    }
    \label{ablation block number}
    \end{table}

    \begin{figure}[ht]
    \centering
    \includegraphics[width=\columnwidth]{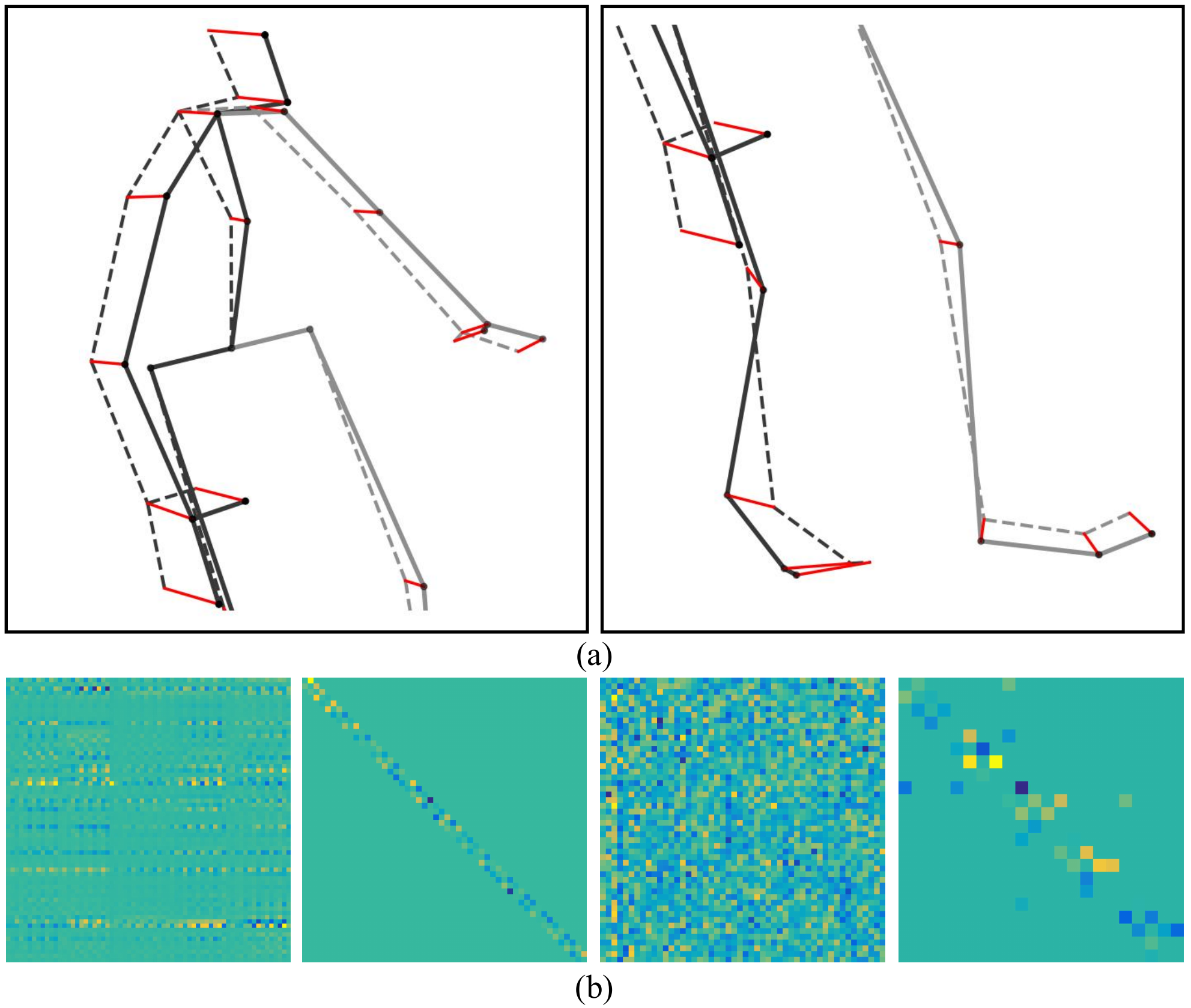} 
    \caption{\textbf{Visualization results.}
    (a) Effect of graph guidance. Solid lines depict the pose learned by the overall network; dashed lines depict the pose learned without graph aggregation, the differences between which are highlighted by red lines, which represent the graph duidance.
    (c) Visualization of the learned spatial Graph-Guided Mixer, spatial adjacency and their temporal counterparts (from left to right).
    }
    \label{viz_upper_lower}
    \end{figure}

\noindent\textbf{Role of the model scale.}
We show the influence of model scaling on our network by decreasing or increasing the number of Graph-Guided Mixer blocks in Table \ref{ablation block number}.
We see that our model achieves its best with 48 blocks.

\subsection{Visualization}
To put a finer point on the evaluation of our method, we present various qualitative results.
Figure \ref{viz} provides samples of walking pose sequences predicted by ours and \cite{guo2023back} on H3.6M, which shows that our method generates more realistic and accurate poses.
To thoroughly analyze the superiority of our network, we show the effect of graph guidance in Figure \ref{viz_upper_lower} (a). The pose represented by solid lines is the final output of the network, while the pose of dashed lines is obtained during inference with aggregation manually set to zeros, thereby eliminating their influence.
These dashed lines demonstrate the structure-agnostic dependencies learned by basic mixers, which are not sufficient for accurate motion prediction. The differences between the two poses, highlighted by the red lines, stress the importance of learning structure-specific dependencies based on adjacency information, as can also be developed from the visualization of the learned network components in Figure \ref{viz_upper_lower} (b). Overall, these results underscore the superiority of our approach.

\subsection{Conclusion}
In this paper, we first develop a theoretical connection between MLP-Mixers and GCNs. We show that they bear a strong analogy in terms of how the individual layer looks, and thus can be combined into an approach that exploits advantages from both sides.
In human motion prediction, we find that both the structure-agnostic and structure-specific information is necessary, because human motion is intrinsically ruled by skeletal structure and shares similarity in motion patterns. However, current approaches largely do not include this information into their networks.
Based on our theoretical findings, we propose Graph-Guided Mixer. By incorporating graph guidance, our Graph-Guided Mixer can effectively capture and utilize the specific connectivity patterns within human skeleton’s graph representation.

\normalem
{\small
\bibliographystyle{ieee_fullname}
\bibliography{main}

\begin{thebibliography}{10}\itemsep=-1pt

\bibitem{bouazizi2022motionmixer}
Arij Bouazizi, Adrian Holzbock, Ulrich Kressel, Klaus Dietmayer, and Vasileios
  Belagiannis.
\newblock Motionmixer: mlp-based 3d human body pose forecasting.
\newblock {\em arXiv preprint arXiv:2207.00499}, 2022.

\bibitem{bruna2013spectral}
Joan Bruna, Wojciech Zaremba, Arthur Szlam, and Yann LeCun.
\newblock Spectral networks and locally connected networks on graphs.
\newblock {\em arXiv preprint arXiv:1312.6203}, 2013.

\bibitem{chen2021cyclemlp}
Shoufa Chen, Enze Xie, Chongjian Ge, Ding Liang, and Ping Luo.
\newblock Cyclemlp: A mlp-like architecture for dense prediction.
\newblock {\em arXiv preprint arXiv:2107.10224}, 2021.

\bibitem{chollet2017xception}
Fran{\c{c}}ois Chollet.
\newblock Xception: Deep learning with depthwise separable convolutions.
\newblock In {\em Proceedings of the IEEE conference on computer vision and
  pattern recognition}, pages 1251--1258, 2017.

\bibitem{cui2020learning}
Qiongjie Cui, Huaijiang Sun, and Fei Yang.
\newblock Learning dynamic relationships for 3d human motion prediction.
\newblock In {\em IEEE/CVF Conference on Computer Vision and Pattern
  Recognition (CVPR)}, pages 6519--6527, 2020.

\bibitem{dang2021msr}
Lingwei Dang, Yongwei Nie, Chengjiang Long, Qing Zhang, and Guiqing Li.
\newblock Msr-gcn: Multi-scale residual graph convolution networks for human
  motion prediction.
\newblock In {\em IEEE/CVF International Conference on Computer Vision (ICCV)},
  pages 11467--11476, 2021.

\bibitem{guo2022hire}
Jianyuan Guo, Yehui Tang, Kai Han, Xinghao Chen, Han Wu, Chao Xu, Chang Xu, and
  Yunhe Wang.
\newblock Hire-mlp: Vision mlp via hierarchical rearrangement.
\newblock In {\em Proceedings of the IEEE/CVF Conference on Computer Vision and
  Pattern Recognition}, pages 826--836, 2022.

\bibitem{guo2023back}
Wen Guo, Yuming Du, Xi Shen, Vincent Lepetit, Xavier Alameda-Pineda, and
  Francesc Moreno-Noguer.
\newblock Back to mlp: A simple baseline for human motion prediction.
\newblock In {\em Proceedings of the IEEE/CVF Winter Conference on Applications
  of Computer Vision}, pages 4809--4819, 2023.

\bibitem{hou2022vision}
Qibin Hou, Zihang Jiang, Li Yuan, Ming-Ming Cheng, Shuicheng Yan, and Jiashi
  Feng.
\newblock Vision permutator: A permutable mlp-like architecture for visual
  recognition.
\newblock {\em IEEE Transactions on Pattern Analysis and Machine Intelligence},
  45(1):1328--1334, 2022.

\bibitem{ionescu2013human3}
Catalin Ionescu, Dragos Papava, Vlad Olaru, and Cristian Sminchisescu.
\newblock Human3.6m: Large scale datasets and predictive methods for 3d human
  sensing in natural environments.
\newblock {\em IEEE Transactions on Pattern Analysis and Machine Intelligence
  (PAMI)}, 36(7):1325--1339, 2013.

\bibitem{kingma2014adam}
Diederik~P Kingma and Jimmy Ba.
\newblock Adam: A method for stochastic optimization.
\newblock {\em ArXiv:1412.6980}, 2014.

\bibitem{kipf2016semi}
Thomas~N Kipf and Max Welling.
\newblock Semi-supervised classification with graph convolutional networks.
\newblock {\em ArXiv:1609.02907}, 2016.

\bibitem{li2018convolutional}
Chen Li, Zhen Zhang, Wee~Sun Lee, and Gim~Hee Lee.
\newblock Convolutional sequence to sequence model for human dynamics.
\newblock In {\em IEEE/CVF Conference on Computer Vision and Pattern
  Recognition (CVPR)}, pages 5226--5234, 2018.

\bibitem{li2021skeleton}
Maosen Li, Siheng Chen, Zihui Liu, Zijing Zhang, Lingxi Xie, Qi Tian, and Ya
  Zhang.
\newblock Skeleton graph scattering networks for 3d skeleton-based human motion
  prediction.
\newblock In {\em Proceedings of the IEEE/CVF international conference on
  computer vision}, pages 854--864, 2021.

\bibitem{li2022skeleton}
Maosen Li, Siheng Chen, Zijing Zhang, Lingxi Xie, Qi Tian, and Ya Zhang.
\newblock Skeleton-parted graph scattering networks for 3d human motion
  prediction.
\newblock In {\em Computer Vision--ECCV 2022: 17th European Conference, Tel
  Aviv, Israel, October 23--27, 2022, Proceedings, Part VI}, pages 18--36.
  Springer, 2022.

\bibitem{li2020dynamic}
Maosen Li, Siheng Chen, Yangheng Zhao, Ya Zhang, Yanfeng Wang, and Qi Tian.
\newblock Dynamic multiscale graph neural networks for 3d skeleton based human
  motion prediction.
\newblock In {\em IEEE/CVF Conference on Computer Vision and Pattern
  Recognition (CVPR)}, pages 214--223, 2020.

\bibitem{li2021multiscale}
Maosen Li, Siheng Chen, Yangheng Zhao, Ya Zhang, Yanfeng Wang, and Qi Tian.
\newblock Multiscale spatio-temporal graph neural networks for 3d
  skeleton-based motion prediction.
\newblock {\em IEEE Transactions on Image Processing (TIP)}, 30:7760--7775,
  2021.

\bibitem{ma2022progressively}
Tiezheng Ma, Yongwei Nie, Chengjiang Long, Qing Zhang, and Guiqing Li.
\newblock Progressively generating better initial guesses towards next stages
  for high-quality human motion prediction.
\newblock In {\em IEEE/CVF Conference on Computer Vision and Pattern
  Recognition (CVPR)}, pages 6437--6446, 2022.

\bibitem{mahmood2019amass}
Naureen Mahmood, Nima Ghorbani, Nikolaus~F Troje, Gerard Pons-Moll, and
  Michael~J Black.
\newblock Amass: Archive of motion capture as surface shapes.
\newblock In {\em Proceedings of the IEEE/CVF international conference on
  computer vision}, pages 5442--5451, 2019.

\bibitem{mao2020history}
Wei Mao, Miaomiao Liu, and Mathieu Salzmann.
\newblock History repeats itself: Human motion prediction via motion attention.
\newblock In {\em Computer Vision--ECCV 2020: 16th European Conference,
  Glasgow, UK, August 23--28, 2020, Proceedings, Part XIV 16}, pages 474--489.
  Springer, 2020.

\bibitem{mao2019learning}
Wei Mao, Miaomiao Liu, Mathieu Salzmann, and Hongdong Li.
\newblock Learning trajectory dependencies for human motion prediction.
\newblock In {\em IEEE/CVF International Conference on Computer Vision (ICCV)},
  pages 9489--9497, 2019.

\bibitem{mao2021multi}
Wei Mao, Miaomiao Liu, Mathieu Salzmann, and Hongdong Li.
\newblock Multi-level motion attention for human motion prediction.
\newblock {\em International journal of computer vision}, 129(9):2513--2535,
  2021.

\bibitem{schwind2020effects}
Valentin Schwind, David Halbhuber, Jakob Fehle, Jonathan Sasse, Andreas
  Pfaffelhuber, Christoph T{\"o}gel, Julian Dietz, and Niels Henze.
\newblock The effects of full-body avatar movement predictions in virtual
  reality using neural networks.
\newblock In {\em Proceedings of the 26th ACM Symposium on Virtual Reality
  Software and Technology}, pages 1--11, 2020.

\bibitem{sofianos2021space}
Theodoros Sofianos, Alessio Sampieri, Luca Franco, and Fabio Galasso.
\newblock Space-time-separable graph convolutional network for pose
  forecasting.
\newblock In {\em IEEE/CVF International Conference on Computer Vision (ICCV)},
  pages 11209--11218, 2021.

\bibitem{tang2022sparse}
Chuanxin Tang, Yucheng Zhao, Guangting Wang, Chong Luo, Wenxuan Xie, and Wenjun
  Zeng.
\newblock Sparse mlp for image recognition: Is self-attention really necessary?
\newblock In {\em Proceedings of the AAAI Conference on Artificial
  Intelligence}, volume~36, pages 2344--2351, 2022.

\bibitem{tang2022image}
Yehui Tang, Kai Han, Jianyuan Guo, Chang Xu, Yanxi Li, Chao Xu, and Yunhe Wang.
\newblock An image patch is a wave: Phase-aware vision mlp.
\newblock In {\em Proceedings of the IEEE/CVF Conference on Computer Vision and
  Pattern Recognition}, pages 10935--10944, 2022.

\bibitem{tolstikhin2021mlp}
Ilya~O Tolstikhin, Neil Houlsby, Alexander Kolesnikov, Lucas Beyer, Xiaohua
  Zhai, Thomas Unterthiner, Jessica Yung, Andreas Steiner, Daniel Keysers,
  Jakob Uszkoreit, et~al.
\newblock Mlp-mixer: An all-mlp architecture for vision.
\newblock {\em Advances in neural information processing systems},
  34:24261--24272, 2021.

\bibitem{touvron2022resmlp}
Hugo Touvron, Piotr Bojanowski, Mathilde Caron, Matthieu Cord, Alaaeldin
  El-Nouby, Edouard Grave, Gautier Izacard, Armand Joulin, Gabriel Synnaeve,
  Jakob Verbeek, et~al.
\newblock Resmlp: Feedforward networks for image classification with
  data-efficient training.
\newblock {\em IEEE Transactions on Pattern Analysis and Machine Intelligence},
  2022.

\bibitem{tu2023consistent}
Zhigang Tu, Zhisheng Huang, Yujin Chen, Di Kang, Linchao Bao, Bisheng Yang, and
  Junsong Yuan.
\newblock Consistent 3d hand reconstruction in video via self-supervised
  learning.
\newblock {\em IEEE Transactions on Pattern Analysis and Machine Intelligence},
  2023.

\bibitem{tu2022general}
Zhigang Tu, Xiangjian Liu, and Xuan Xiao.
\newblock A general dynamic knowledge distillation method for visual analytics.
\newblock {\em IEEE Transactions on Image Processing}, 31:6517--6531, 2022.

\bibitem{tu2023dtcm}
Zhigang Tu, Yuanzhong Liu, Yan Zhang, Qizi Mu, and Junsong Yuan.
\newblock Dtcm: Joint optimization of dark enhancement and action recognition
  in videos.
\newblock {\em IEEE Transactions on Image Processing}, 2023.

\bibitem{tu2022maxim}
Zhengzhong Tu, Hossein Talebi, Han Zhang, Feng Yang, Peyman Milanfar, Alan
  Bovik, and Yinxiao Li.
\newblock Maxim: Multi-axis mlp for image processing.
\newblock In {\em Proceedings of the IEEE/CVF Conference on Computer Vision and
  Pattern Recognition}, pages 5769--5780, 2022.

\bibitem{unhelkar2018human}
Vaibhav~V Unhelkar, Przemyslaw~A Lasota, Quirin Tyroller, Rares-Darius Buhai,
  Laurie Marceau, Barbara Deml, and Julie~A Shah.
\newblock Human-aware robotic assistant for collaborative assembly: Integrating
  human motion prediction with planning in time.
\newblock {\em IEEE Robotics and Automation Letters}, 3(3):2394--2401, 2018.

\bibitem{vaswani2017attention}
Ashish Vaswani, Noam Shazeer, Niki Parmar, Jakob Uszkoreit, Llion Jones,
  Aidan~N Gomez, {\L}ukasz Kaiser, and Illia Polosukhin.
\newblock Attention is all you need.
\newblock {\em Advances in neural information processing systems}, 30, 2017.

\bibitem{von2018recovering}
Timo Von~Marcard, Roberto Henschel, Michael~J Black, Bodo Rosenhahn, and Gerard
  Pons-Moll.
\newblock Recovering accurate 3d human pose in the wild using imus and a moving
  camera.
\newblock In {\em Proceedings of the European conference on computer vision
  (ECCV)}, pages 601--617, 2018.

\bibitem{yu2022s2}
Tan Yu, Xu Li, Yunfeng Cai, Mingming Sun, and Ping Li.
\newblock S2-mlp: Spatial-shift mlp architecture for vision.
\newblock In {\em Proceedings of the IEEE/CVF Winter Conference on Applications
  of Computer Vision}, pages 297--306, 2022.

\bibitem{zhang2012imageadmixture}
Fang-Lue Zhang, Ming-Ming Cheng, Jiaya Jia, and Shi-Min Hu.
\newblock Imageadmixture: Putting together dissimilar objects from groups.
\newblock {\em IEEE Transactions on Visualization and Computer Graphics},
  18(11):1849--1857, 2012.

\bibitem{zhang2018detecting}
Fang-Lue Zhang, Xian Wu, Rui-Long Li, Jue Wang, Zhao-Heng Zheng, and Shi-Min
  Hu.
\newblock Detecting and removing visual distractors for video aesthetic
  enhancement.
\newblock {\em IEEE Transactions on Multimedia}, 20(8):1987--1999, 2018.

\bibitem{zhong2022spatio}
Chongyang Zhong, Lei Hu, Zihao Zhang, Yongjing Ye, and Shihong Xia.
\newblock Spatio-temporal gating-adjacency gcn for human motion prediction.
\newblock In {\em Proceedings of the IEEE/CVF Conference on Computer Vision and
  Pattern Recognition}, pages 6447--6456, 2022.

\bibitem{zhou2021learning}
Honghong Zhou, Caili Guo, Hao Zhang, and Yanjun Wang.
\newblock Learning multiscale correlations for human motion prediction.
\newblock In {\em IEEE International Conference on Development and Learning
  (ICDL)}, pages 1--7, 2021.

\end{thebibliography}
}

\end{document}